\documentclass [a4paper] {article}
\usepackage[width=185mm,top=15mm,bottom=15mm]{geometry}
\usepackage[utf8]{inputenc}
\usepackage{graphicx}
\usepackage{float}
\usepackage{amssymb}
\usepackage{amsmath}
\usepackage{enumerate}
\usepackage{color}
\usepackage{tikz}
\usepackage{tabularx}
\usepackage{hyperref}
\title{\bf{Fuzzy Logic Controller Design for Mobile Robot Outdoor Navigation  }}

\author{ \\ Assefinew Wondosen* and Dereje Shiferaw**\\ \\ \small{*Adama Sceince and Technology University, \textcolor{blue}{wondebly@gmail.com}} \\ \small{**Addis Ababa Institute of Technology, \textcolor{blue}{ dnderejesh@gmail.com}}\\} 

\date{April 2017}
\begin{document}
\maketitle
\begin{abstract}
Many researchers around the world are researching to get control solutions that enhance robots' ability to navigate in dynamic environments autonomously. However, until these days robots have limited capability and many navigation tasks on Earth and other planets have been difficult so far. \par
This paperwork presents the development of a control system for a differential drive-wheeled mobile robot that autonomously controls its position, heading, and speed based on destination information given and surrounding data gathered through mounted proximity and GPS sensors. The intelligence of this control system is implemented by using a fuzzy logic algorithm which is a very powerful tool to handle un-modeled systems like the dynamically changing environment dealt with in this research. The fuzzy controller is used to address the problems associated with navigation in an obstacle-strewn environment. Such issues include position estimation, path planning, and obstacle avoidance.  \par
In this study modeling, design, and simulation of the system have been done. The simulation result shows that the developed mobile robot travels successfully from any location to the destination location without colliding with obstacles.  \par

\end{abstract}
\href{https://github.com/WondesenB/FuzzyMobileRobotGuidance.git}{{https://github.com/WondesenB/FuzzyMobileRobotGuidance.git}}
\section{Introduction}
Currently, research in autonomous mobile robots gained an extensive interest. This is due to the wish to replace human with robots in dangerous tasks and to use them in everyday life tasks and industry. Different types of mobile robots are under research and used for different purposes. These are wheeled, legged, aerial, underwater, and humanoid robots \cite{1}. In this paper wheeled mobile robot outdoor navigation challenge is chosen for study.\par
Mobile robots are Electro-mechanical devices capable of moving in an environment with a certain degree of autonomy [15]. Autonomous navigation is assisted by different information-gathering sensors. The most common sensors are distance sensors (ultrasonic, laser, etc.) capable of detecting obstacles closer to the robot's path [15]. The other is the Global Positioning System (GPS) sensor which identifies the robot's location.  Knowledge of self-location, and the location of other places of interest in the world, is the basic foundation on which all high-level navigation operations are built. Therefore, knowing the location enables strategic path planning for tasks such as goal searching, region coverage, exploration, and obstacle avoidance, and makes following these planned trajectories possible [13]. All these tasks are done by the control system deployed on the robot. \par
  Following the great advancement of electronics and computer technology, control system mechanisms have developed a lot and now reached the era of artificial intelligence (AI) that can handle nonlinear situations. This is due to the emergence of neural networks, fuzzy logic, genetic algorithms,s and other soft computing algorithms. Due to its simplicity and capability for real-time implementation, the fuzzy logic controller is found to be a very convenient controller for systems that have no mathematical model. Fuzzy logic works based on rules that are a direct reflection of the way a human being uses its brain to accomplish different tasks. From the perspective of the complexity of navigation tasks and fuzzy logic's suitability to handle such problems, in this paper fuzzy logic controller is chosen as a controller for mobile robot navigation to a goal position in an unknown environment.

\subsection{Fuzzy Logic}
Fuzzy logic is an algorithm used to produce decisions based on input conditions. Fuzzy logic involves the use of qualitative reasoning instead of purely quantitative measurements. It is particularly useful when dealing with systems that are ill-defined or difficult to model. Like Boolean logic, fuzzy logic involves selecting an action if a set of conditions are satisfied. However, rather than deciding whether the conditions are true or false, fuzzy logic estimates each condition’s degree of truth (generally a number between 0 and 1), and calculates output values based on the relative estimates [14]. The fuzzy controller has four main components [2]. \par
\begin{enumerate}
\item	\textbf{Rule-base:} holds the knowledge, in the form of a set of rules, of how best to control the system.
\item	\textbf{Inference mechanism:} evaluates which control rules are relevant at the current time and then decides what the input to the plant should be. 
\item  \textbf{Fuzzification:} simply modifies the inputs so that they can be interpreted and compared to the rules in the rule base. 
\item 	\textbf{Defuzzification:} converts the conclusions reached by the inference mechanism into the inputs to the plant \cite{2}.
\end{enumerate}

The fuzzy approach is especially successful in the case of processes such as washing machines, robot control, camera focusing, or drying processes that are being controlled well by human operators \cite{8}.

\section{Literature Review}
Zadeh's contribution to mathematical logic models led to a wave of research based on his concept of 'fuzzy logic'. An area of research being explored with his notion of fuzzy sets is the design and implementation of intelligent control systems termed Fuzzy Logic (FL). With only a short period since the concept's introduction, explorations into control design have been an area of heavy interest. Current research with this controller revolves around experimenting with different applications as well as speeding the design of implementation \cite{2}.\par
In \cite{15} an autonomous wheeled wall-following robot was designed using ultrasonic sensors for inputs to traverse a known indoor environment for an IEEE competition. The FL controller for the robot was designed using MATLAB for simulation and utilization of a PIC microcontroller for implementation. The differential drive of the robot was developed based on the kinematic equation. The FL controller has two inputs: the position error, and the angle error. These values are gathered from the three mounted ultrasonic sensors on the front and two sides of the robot base. The controller employs the use of 18 rules to process the fuzzy data. Through defuzzification, two outputs are generated for position correction and angle correction sent to the servo motors. A barrier to overcome by research while implementing the controller was a processing time issue. The system clock of the microcontroller was 4MHz, which translated into a 0.4-second processing time from fuzzification, rule processing, and defuzzification. To bypass this issue, the pair generated a look-up table to load onto the microcontroller in place of the FL controller. The results included an efficiently performing controller able to reach a referenced wall distance from any angle starting position. \par
In another study by Y. Ino et. \cite{24} a controller for testing a mobile robot base for corridor navigation has been designed and created. The four units responsible for different aspects of the control system were used. These are sensor handling, machine vision, collision avoidance using FL, and locomotion. The infrared sensors are spaced to provide 360-degree coverage. The fuzzy collision avoidance portion of the controller utilizes one input fuzzy set for the sensor inputs and three output fuzzy sets. Seventeen rules are employed by the rule processing section of the FL controller. The outputs include distance, velocity, and turn angle. The distance output membership function (MF) determines if the output should move the robot forward or backward. The velocity output MF uses the linguistic variables slow, medium, and fast to generate an appropriate speed. The turn angle output MF divides the total angle, pre-set to 60 degrees, into sections of positive left, negative left, positive center, negative center, positive right, and negative right. The result was a mobile robot that avoided collisions with both obstacles and walls in a real indoor environment. The FL controller produced an unwanted zig-zag path pattern sometimes. It was also determined that infrared sensors were negatively affected by the ambient indoor lighting.\par
In \cite{33} fuzzy approach for solving the motion planning problem of a mobile robot in the presence of moving obstacles was derived. This consists of formulating a general method for the derivation of input-output data to construct a fuzzy logic controller (FLC) line. The FLC is constructed based on the use of a recently developed data-driven and fuzzy controller modeling algorithm and it can then be used online by the robot to navigate among moving obstacles. The novelty of the approach, as compared to the most recent fuzzy ones, stems from its generality. The formulated data-derivation method enables the construction of a single FLC to accommodate a wide range of circumstances. \par
In \cite{28} fuzzy logic controller has been designed for real real-time mobile robot navigation system in an unknown environment. To fulfill the optimization requirement this study considered the optimization of the time as well as the distance traveled by the robot. The fuzzy logic navigation technique developed in this work was implemented on a real robot by using seven ultrasonic sensors to perceive the environment. However, this paper didn’t try to explain the methods or devices used to identify the robot’s absolute or relative location on Earth. The robot is also limited to responding to static obstacles only. The Simulation and experimental result show that the robot able to reach the goal position safely.\par
In \cite{29} fuzzy controller path planning for mobile robots was presented. In this paper obstacle detection was implemented by using 24 Ultrasonic sensors in a gap distance of $15^o$ around the robot. Increasing the number of obstacle-detecting ultrasonic sensors installed around the robot has a great advantage in increasing the obstacle-detection capability. However, this could lead to an increase in cost and high processor speed requirements.  \par
In \cite{30} Object Based Navigation of Mobile Robots with Obstacle Avoidance using a Fuzzy logic Controller was presented. The focus of this paper has been on developing autonomous urban search and rescue robots. The search and rescue robot is equipped with a victim detection system which determines the robot's target.  In this paper, the navigation approach has been extended using a fuzzy controller that will take a path based on extracted lines and fuse data from sonar modules. By data fusing technique robot can extract a crystalline object that are hidden from the laser and can avoid dynamically placed obstacles near and along the path. It is the nice approach of this paper that the problem of laser scanners on crystalline objects like glass and mirrors is solved by using data fusing from sonar sensors.  Furthermore, it has good real-time capability and it is implemented on the NAJI V mobile search and rescue robot platform. However, the applicability of the design is limited to small areas like office rooms. \par
In \cite{31} a fuzzy logic-based model is presented for the navigation of mobile robots in indoor environments. The inputs of the fuzzy controller are the outputs from the sensor system, including the obstacle distances obtained from the left, front, and right sensor groups, the target direction, and the current robot speed. A set of linguistic fuzzy rules are developed to implement expert knowledge under various situations. The output signals from the fuzzy controller are the accelerations of the left and right wheels, respectively. Under the proposed Fuzzy model, a mobile robot avoids the obstacles and generates the path towards the target. The design approach in this paper is good; however, the proposed fuzzy controller for the mobile robot is for indoor navigation. This couldn’t be used for outdoor navigation.\par
In \cite{32} A New Fuzzy Intelligent Obstacle Avoidance Control Strategy for Wheeled mobile Robot proposed in 2012. This paper described a fuzzy intelligent obstacle avoidance control strategy for wheeled mobile robots in changing environments with obstacles, which utilizes fuzzy control methodology imitating human driving intelligence and has the merit of fast obstacle avoidance and quick response relative to traditional model-based methods. Additionally, this paper proposed an intelligent coordinator that coordinates outputs of run-to-goal fuzzy controller and fuzzy obstacle avoidance controller effectively. In this paper, the localization techniques and the mathematical tools used to identify the robot's absolute and relative position are not discussed. The localization issue is equally important as the controller.\par
Research on mobile robot control is continuing. Some of the research mentioned in this section focuses on designing of fuzzy controller for indoor navigation, wall following, line following, and path planning by avoiding obstacles. In this research, the mobile robot considered is for use in outdoor navigation in a dynamic environment. The controller is based on a fuzzy logic algorithm. In addition, the localization techniques used in this research are a hybrid type. One is based on GPS sensor position reading and the other is by position estimation based on wheel speed encoder reading. 
\section{Mobile Robot Motion Modeling}
The steering of a mobile robot is achieved by differentially driven wheel pairs on each side of the robot. From a control viewpoint, the unusual nature of nonholonomic kinematics and the dynamic complexity of the mobile robot make feedback stabilization at a given posture difficult to achieve via smooth time-invariant control. This indicates that the problem is truly nonlinear; linear control is ineffective and innovative design techniques are needed. It is an accepted practice to work with dynamical models to obtain stable motion control laws for trajectory following or goal-reaching. The robot has two driving wheels mounted on the same axis and free front wheels. Two driving wheels are independently driven by two actuators to achieve both the transition and orientation. The position of the mobile robot in the global frame {Latitude, Longitude} can be defined by the position of the mass center of the mobile robot system.\par
The mathematical model of a mobile robot moving on a planner surface can be decomposed into three sub-systems which are the kinematic model, dynamic model, and electric drive system. This approach is quite natural for most mobile robots powered by electric motors \cite{10}.

\begin{figure}[H]
\centering
\begin{tikzpicture}[thick,scale=1.25]
\draw [->,color=orange,very thick](0,1)--(1,1);
\draw(-0.2,1.25)node {{Electrical}};
\draw(-0.2,0.75)node {{Input}};

\draw [very thick,color=red](1,0) rectangle (3,2);
\draw(2,1.5)node {{Electrical }};
\draw(2,1)node {{Drive }};
\draw(2,0.5)node {{Subsystem}};

\draw (3.75,1.25) node {{Torque}};
\draw [->,color=violet, very thick](3,1)--(4.5,1);
\draw [very thick,color=blue](4.5,0) rectangle (6.5,2);
\draw(5.5,1.25)node {{Dynamic }};
\draw(5.5,0.75)node {{Subsystem }};
 
\draw (7.25,1.25) node {{Velocity}};
\draw [->,color=orange,very thick](6.5,1)--(8,1);
\draw [very thick,color=green](8,0) rectangle (10,2);
\draw(9,1.25)node {{Kinematic }};
\draw(9,0.75)node {{Subsystem }};

\draw [->,color=yellow, very thick](10,1)--(11,1);
\draw (11.3,1.25) node {{Position}};
\draw[dashed](0.5,2.5) -- (10.5,2.5);
\draw[dashed](10.5,2.5) -- (10.5,-0.5);
\draw[dashed](0.5,-0.5) -- (10.5,-0.5);
\draw[dashed](0.5,-0.5) -- (0.5,2.5);
\end{tikzpicture}
\caption{Decomposition of an electrically driven Mobile Robot}
\end{figure}
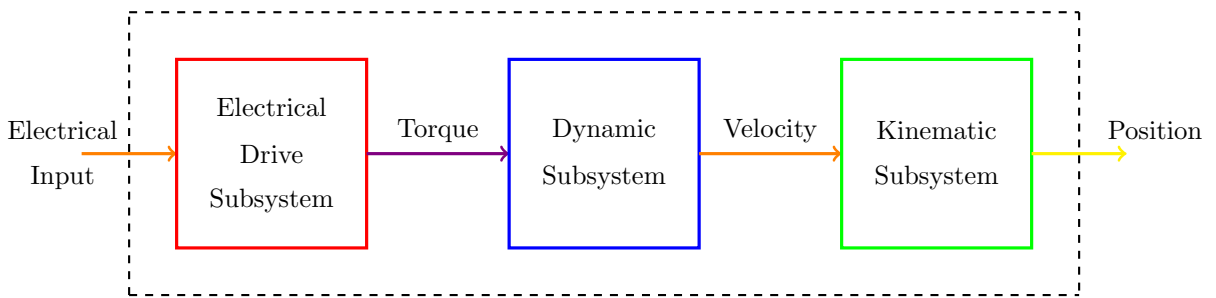
 \subsection{Kinematic Model }
 The robot has two identical parallel, non-deformable rear wheels which are controlled by two independent motors, and steering front wheels. It is assumed that the plane of each wheel is perpendicular to the ground and the contact between the wheels and the ground is pure rolling and non-slipping, i.e., the velocity of the center of mass of the robot is orthogonal to the rear wheels’ axis. The center of mass of the mobile robot is located in the middle of the axis connecting the rear wheels.
As we can see in the following diagram, $\{X, Y\}$ is the reference frame while $\{X_m, Y_m\}$ is the moving mobile robot reference frame. The equation of mobile robot motion can be obtained as follows. 
\begin{figure}[H]
\centering
\begin{tikzpicture}[scale=1.5]
\draw [->](0,0)--(0,3);
\draw [->,very thick,color=red](2,0)--(2,1.1547);
\draw [->](0,0)--(3,0);
\draw [->,very thick,color=red](0,0)--(2,0);
\draw (1.75,-0.25)node {\small{$\dot{X}=v_x(t)cos\theta$}};
\draw (3.2,0.5)node {\small{$\dot{Y}=v_x(t)sin\theta$}};
\draw[->,dashed](0,0)--(3,1.732);
\draw[->,very thick](0,0)--(2,1.1547);
\draw (1.5,1.25)node [rotate=30]{\small{$v_x(t)$}};
\draw[->,dashed](0,0)--(-1.732,3);
\draw[->,very thick](0,0)--(-1.1547,2);
\draw (-1.5,1.7)node [rotate=120]{\small{$v_y(t)$}};
\draw [->](1,0) arc (0:30:1);
\draw (1.3,0.25)node {\small{$\theta$}};
\draw (-0.3,1.25)node {\small{$\theta$}};
\draw [->](0,1) arc (90:120:1);
\end{tikzpicture}
\caption{Robot free body diagram}
\end{figure}
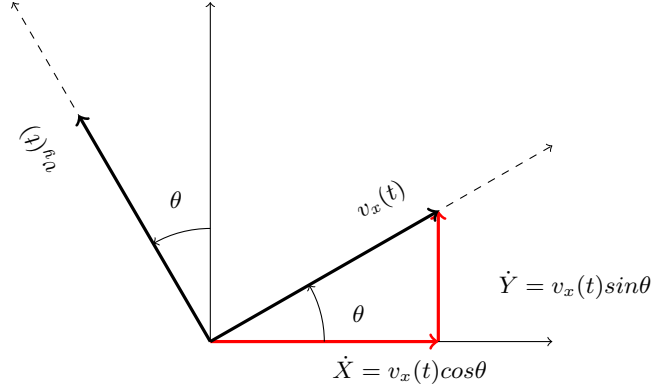

Suppose that the robot moves on a plane with linear velocity expressed in the reference frame as $V = {[ V_x \quad  V_y \quad  0]} ^T $ and rotates with an angular velocity vector $\dot{\theta} =  {[ 0\quad 0 \quad \dot{\theta} ]}^T$. If $q = {[ X \quad Y \quad \theta]}^T$ is the state vector describing generalized coordinates of the robot (i.e., the center of mass position, x and y, and the orientation  $\theta$ of the reference coordinate frame concerning the inertial frame), then $\dot{q}= {[\dot{X}\quad \dot{Y} \quad \dot{\theta}]}^T$ denotes the vector of generalized velocities.\vspace{0.5cm}
\begin{figure}[H]
\centering
\begin{tabular}{m{6cm}m{6cm}}
\begin{tikzpicture}[scale=1.25]
\begin{scope}[scale=0.9]
\draw [->](0,0)--(0,3);
\draw [->](0,0)--(3,0);
\draw [rotate=45](1.5,-1) rectangle (3.5,0.5);
\draw [fill=gray,rotate=45](1.7,-1.15) rectangle (2.1,-1);
\draw [fill=gray,rotate=45](3,-1.15) rectangle (3.4,-1);
\draw [fill=gray,rotate=45](1.7,0.5) rectangle (2.1,0.65);
\draw [fill=gray,rotate=45](3,0.5) rectangle (3.4,0.65);
\draw [->,thick](2,1.75)--(1,2.75);
\draw (1.0,2.25)node [rotate=45]{\tiny{$V_y(t)$}};
\draw [->,dashed](2,1.75)--(0.5,3.25);
\draw [->,thick](2.0,1.75)--(3,2.75);
\draw (3.25,2.5)node [rotate=45]{\tiny{$V_x(t)$}};
\draw [->,dashed](2.0,1.75)--(3.25,3);
\draw[dashed](2,0)--(2,1.75);
\draw[dashed](0,1.75)--(3.5,1.75);
\draw (2.5,1.75) arc (0:45:0.5);
\draw (2.7,2)node {\small{$\theta$}};
\draw (-0.25,3)node{\small{Y}};
\draw (3.25,0)node{\small{X}};
\draw (2.75,2.85)node[rotate=45]{\tiny{$X_m$}};
\draw (0.6,2.85)node[rotate=135]{\tiny{$Y_m$}};
\draw (2,-0.1)node{\tiny{$x$}};
\draw (-0.2,1.75)node{\tiny{$y$}};
\end{scope}
\end{tikzpicture} &

\begin{tikzpicture}[scale=1.25]
\begin{scope}[scale=0.9]
\draw [->](-1.5,0)--(-1.5,3);
\draw [->](-1.5,0)--(3,0);
\draw [rotate=45](1.5,-1) rectangle (3.5,0.5);
\draw [fill=gray,rotate=45](1.7,-1.15) rectangle (2.1,-1);
\draw [fill=gray,rotate=45](3,-1.15) rectangle (3.4,-1);
\draw [fill=gray,rotate=45](1.7,0.5) rectangle (2.1,0.65);
\draw [fill=gray,rotate=45](3,0.5) rectangle (3.4,0.65);

\draw [->,thick](2,1.75)--(1,2.75);
\draw (1.0,2.25)node [rotate=45]{\tiny{$V_y(t)$}};
\draw [->,dashed](2,1.75)--(0.5,3.25);
\draw [->,thick](2.0,1.75)--(3,2.75);
\draw (3.25,2.5)node [rotate=45]{\tiny{$V_x(t)$}};
\draw [->,dashed](2.0,1.75)--(3.25,3);
\draw[dashed](2,0)--(2,1.75);
\draw[dashed](-1.5,1.75)--(3.5,1.75);
\draw (2.5,1.75) arc (0:45:0.5);
\draw (2.7,2)node {\small{$\theta$}};
\draw (-1.6,3)node{\small{Y}};
\draw (3.25,0)node{\small{X}};
\draw (2,-0.1)node{\tiny{$x$}};
\draw (-1.6,1.75)node{\tiny{$y$}};
\draw (-0.25,3)circle (1mm);
\draw [<->](2.2,0.5)--(-0.25,3);
\draw [<->](1,0.75)--(-0.75,2.5);
\draw[dashed](2,1.75)--(0.5,0.25);
\draw (-0.75,2.9)node {\small{ICC}};
\draw (-0.0,1.25)node {\small{R}};
\draw (1.5,1) node [rotate=135]{\small{\textsl{l}}};
\end{scope}
\end{tikzpicture}
\end{tabular}
\end{figure}
Where, 
\begin{itemize}
        \item l  is the length between the two rear wheels' axis
	\item ICC  is the instantaneous center of curvature
	\item R  is the instantaneous curvature radius of the robot trajectory, relative to the mid-axis
	\item r  is the wheel radius
\end{itemize}
	
Here the translational and rotational velocities of the mobile robot depend on the linear velocity of the left wheel $(v_l(t))$ and right wheel $(v_r(t))$. 
$$ICC = (x-R sin\theta, y+R cos\theta)$$
$R - \frac{l}{2}$  Curvature radius of trajectory described by left wheel.
$R + \frac{l}{2}$  Curvature radius of trajectory described by Right wheel. Then the angular velocity of the mobile robot on the ICC axis of rotation can be expressed as:
$$ $$
\begin{equation}
\dot{\theta}(t) = \frac{v_l(t)}{R-\frac{l}{2}} =\frac{v_r(t)}{R+\frac{l}{2}}
\end{equation}
\begin{equation}
\label{eq_thetadot}
\dot{\theta}(t) = \frac{v_r(t)-v_l(t)}{l}
\end{equation}
using eq.( \ref{eq_thetadot}) the radius of curvature can be expressed as follows.
\begin{equation}
R=\frac{1}{2} \frac{(v_r(t) + v_l(t))}{(v_l(t)-v_r(t))}
\end{equation}
Then the instantaneous tangential velocity to the radius of curvature will be:
\begin{equation}
\label{eq34}
v(t) = \dot{\theta}(t)R = \frac{1}{2}((v_r(t)+v_l(t))
\end{equation}
Since there is no lateral skidding, the mobile robot has velocity along its x-axis only. The velocity in the y  direction is zero. Then $v(t)$  in eq. \ref{eq34} is equivalent to $V_x(t)$. The kinematic model of mobile robot on its frame of reference then becomes.
\begin{equation}
\begin{bmatrix}
v_x(t) \\ v_y(t) \\ \dot{\theta}(t)
\end{bmatrix} = 
\begin{bmatrix}
\frac{r}{2} & \frac{r}{2} \\ 0 & 0 \\ \frac{-r}{l} & \frac{-r}{l}
\end{bmatrix}
\begin{bmatrix}
\omega_l(t) \\ \omega_r(t)
\end{bmatrix}
\end{equation}
The position of robot can be expressed as follows
\begin{equation}
\begin{bmatrix}
X\\Y\\\theta
\end{bmatrix} = \begin{bmatrix}
x +R(sin(\theta+\omega \Delta t) -sin(\theta)) \\
x + R(-cos(\theta + \omega \Delta t) +cos(\theta))\\
\theta + \omega \Delta t
\end{bmatrix}
\end{equation}
\subsection{Dynamic Model}
A common actuator in control systems is the DC motor. It directly provides rotary motion and, coupled with wheels, can provide transitional motion. The electric circuit of the armature and the free-body diagram of the rotor are shown in the following figure:
\begin{figure}[H]
\centering
\label{fig35}
\centering
\includegraphics[width=0.8\textwidth]{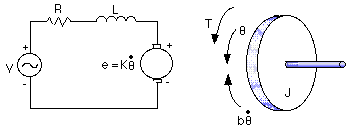} 
\caption{Electric circuit model of DC motor}
\end{figure}
The motor torque (T) is related to the armature current (i) by a constant factor $K_t$. The back emf (e) is also related to the rotational velocity by a constant factor $K_e$. These two parameters are described by the following equations:
\begin{equation}
T=K_ti
\end{equation}
\begin{equation}
e=K_e\frac{d\theta}{dt}
\end{equation}
From fig. 4 we can write the following equations based on Newton's law combined with Kirchhoff's law:
\begin{equation}
\label{eq310}
J\ddot{\theta}+ b\dot{\theta} = K_t I
\end{equation}
\begin{equation}
\label{eq311}
L\frac{di}{dt}+ Ri = V- K_e \dot{\theta}
\end{equation}
Where J is a moment of inertia of the rotor $(Kg.m^2/s^2)$, b is the damping ratio of the mechanical system (Nms), L is inductance (H), R is resistance (ohm) and V is source voltage (V).

The state space model of DC motor can be obtained from eq (\ref{eq310}) \& (\ref{eq311}) as follows.
\begin{align}
\frac{d}{t}\begin{bmatrix}
\dot{\theta} \\ i
\end{bmatrix}& = \begin{bmatrix}
-\frac{b}{J} & \frac{K_t}{J} \\ -\frac{K_e}{L} & - \frac{R}{L}
\end{bmatrix} \begin{bmatrix}
\dot{\theta} \\ i
\end{bmatrix} + \begin{bmatrix}
0 \\ \frac{1}{L}
\end{bmatrix} V  \\
\dot{\theta}& = \begin{bmatrix}
1 & 0\end{bmatrix} \begin{bmatrix}
\dot{\theta} \\ i
\end{bmatrix}
\end{align}

\section{Localization and Path Planning}
In developed engineering and technology, the concept of autonomy of mobile robots covers many areas of knowledge, methodologies, and ultimately algorithms designed for trajectory control, obstacle avoidance, localization, map building, and so on. Practically, the success of a path planning and navigation mission of an autonomous mobile robot depends on the availability of an accurate representation of the navigation environment \cite{22}.\par
In the case of a mobile robot, the specific aspect of cognition is directly linked to robust mobility in navigation competence. Given partial knowledge about its environment and a goal position or series of positions, navigation includes the ability of the robot to act based on its knowledge and sensor values to reach its goal positions as efficiently and as reliably as possible \cite{1}.\par
Self-localization and path planning are two important issues for autonomous mobile robot navigation. For the autonomous motion, the robot should have the ability to localize its current position, perform path planning for future movement, and move to the next position as expected. Currently, there exist many techniques for mobile robot localization. The approaches can be categorized into two groups based on the property of the adopted sensor for data acquisition. One technique that uses internal sensors such as an odometer, gyro, and accelerometer is called dead reckoning. The other adopts external sensors such as ultrasound, laser scanner or camera systems, GPS, etc. This approach relies on the information obtained from the environment and can avoid the internal error made by the robot motion. To increase the applicability of the sensor information, many researchers have combined various types of sensors to achieve better localization results \cite{20}. In this work, GPS, odometer, and ultrasonic sensors are used for the mobile robot localization.\par
To address mobile robot navigational problems, several efficient path-planning techniques have been developed by many researchers. This research will present a detailed analysis of various techniques used in autonomous mobile robot navigation.
\subsection{Mobile Robot Localization}
Navigation is one of the most challenging competencies required of a mobile robot. Success in navigation requires success at the four building blocks of navigation: perception, the robot must interpret its sensors to extract meaningful data; localization, the robot must determine its position in the environment; cognition, the robot must decide how to act to achieve its goals; and motion control, the robot must modulate its motor outputs to achieve the desired trajectory. \par
If one could attach an accurate GPS (global positioning system) sensor to a mobile robot, much of the localization problem would be obviated. The GPS would inform the robot of its exact position, indoors and outdoors, so that the answer to the question, “Where am I?” would always be immediately available. Unfortunately, such a sensor is not currently practical. The existing GPS network provides accuracy to within several meters, which is unacceptable for localizing human-scale mobile robots as well as miniature mobile robots such as desk robots and the body-navigating Nano robots of the future. Furthermore, GPS technologies cannot function indoors or in obstructed areas and are thus limited in their workspace. But, looking beyond the limitations of GPS, localization implies more than knowing one’s absolute position in the Earth’s reference frame. Consider a robot that is interacting with humans. This robot may need to identify its absolute position, but its relative position concerning the target is equally important. Its localization task can include identifying the surroundings using its sensor array and then computing its relative position to the obstacles. Furthermore, during the cognition step a robot will select a strategy for achieving its goals. If it intends to reach a particular location, then localization may not be enough. The robot may need to acquire or build an environmental model, a map that aids it in planning a path to the goal. Once again, localization means more than simply determining an absolute pose in space; it means building a map, and then identifying the robot’s position relative to that map.\par
 Clearly, the robot’s sensors and effectors play an integral role in all the above forms of localization. It is because of the inaccuracy and incompleteness of these sensors and effectors that localization poses difficult challenges \cite{1}.
\begin{figure}[H]
\centering
\label{fig51}
\includegraphics[width=0.8\textwidth]{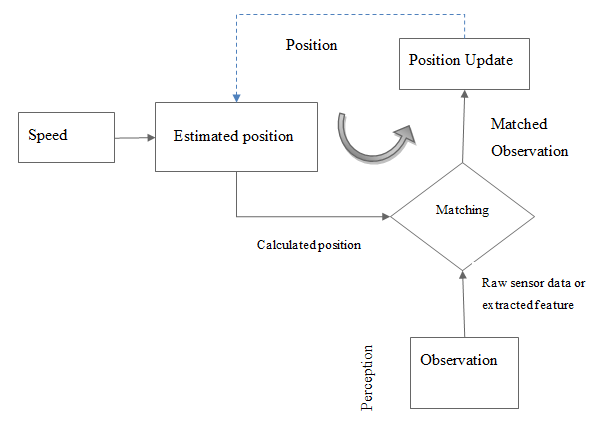} 
\caption{Mobile Robot Localization}
\end{figure}
\subsection{Path Planning}
Given a map and a target location, path planning involves identifying a trajectory that will cause the robot to reach the target location when executed. In other words, path planning is a strategic problem-solving competence, as the robot must decide what to do over the long term to achieve its goals.
The other competence is equally important but occupies the opposite, tactical extreme. Given real-time sensor readings, obstacle avoidance means modulating the trajectory of the robot to avoid collisions. A great variety of approaches have demonstrated competent obstacle avoidance, and we survey a number of these approaches as well \cite{1}.\par
The first step of any path-planning system is to transform this possibly continuous environmental model into a discrete map suitable for the chosen path-planning algorithm. Path planners differ as to how they affect this discrete decomposition. We can identify many general strategies for decomposition; \par
Due to the multitude of path-planning techniques developed so far, one must choose to adapt the path-planning method to the application. The developer will choose between these techniques based on different criteria. These criteria are speed of operation, ability to find the shortest path, and applicability to the scale of obstacles and complexity of the environment. It is found that the heuristic approaches (Fuzzy logic, Neural Network, and Ant Colony Optimization) give suitable and effective results for mobile robot navigation (target reaching and obstacle avoidance) in an unknown and dynamic environment compared to classical techniques. However, based on the processor computational speed requirement relatively fuzzy logic can be implemented on slow processors.
\section{Controller Design}
There are different types of control systems currently used in various application areas. The most commonly used control system today in the industry is the PID controller. In this research fuzzy control system will be used to address mobile robot navigation problems. In fuzzy controller design, the first step is determining the inputs to the controller and the outputs to be controlled.
\par
In navigation of an unknown environment, different types of information about the surrounding area are required. The first one is, knowing where the robot is located at a time. To be able to know this, a GPS sensor is used. Then to get to the destination, in which direction should the robot move? For this purpose, a compass is necessary to align the robot to the direction of its destination. Rotating the robot in the proper direction also needs some kind of mechanization. Therefore, from this sensory information, we do have two inputs, distance to the destination and direction.\par
The second step is, knowing how to get to the destination. How to identify the path that is obstacle-free or less risky? The sensors that help to know about this information are obstacle-detecting sensors (ultrasonic sensors). These sensors will be mounted to the four sides of the robot. Then the sensors measure the distance from the robot to the obstacle and this is the information that is going to indicate whether there is an obstacle or not. Then the following inputs are identified as a very essential inputs to the controller to make decisions on the robot's motion. \\
These inputs are:
\begin{enumerate}[i)]
\item Distance to the destination
\item Direction of destination from current location
\item Front obstacle 
\item Back obstacle
\item Left obstacle
\item Right obstacle
\end{enumerate}
Why is so important to know about the surroundings and destination? What is going to be done with this information?  In the previous section, the inputs to the controller are identified. Now what we want the fuzzy controller to do is, to make the robot move in a way it will arrive at the destination. To achieve this task, different types of locomotion mechanisms are used. Here in this research differential wheel drive method is used. The robot has two front freely moving wheels and two back independently controlled wheels. The robot will move with the desired speed and direction by adjusting the speed of the two back wheels. DC motor is chosen to drive the back wheel because of its simplicity of control. Therefore, the outputs to be controlled are the speed of DC motors that are directly coupled to the right and left back wheels of the robot. The ultimate goal is to decrease the relative distance from the robot to the final destination to zero.
\begin{figure}[H]
\centering
\includegraphics[width=0.8\textwidth]{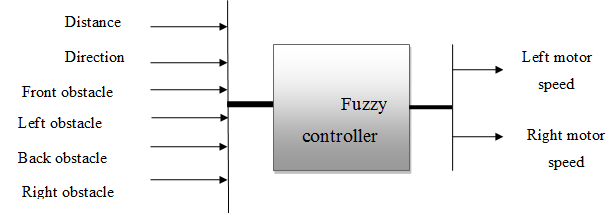} 
\caption{Fuzzy controller outputs}
\end{figure}
\subsection{Fuzzy Controller Design}
The first step in designing a fuzzy controller is the fuzzification of the inputs and outputs. This includes the representation of the range of values with linguistic values. Here in representing linguistic values, we should be careful in choosing the number of linguistic values, the more the number of linguistic values, the more the controller will be accurate and the more powerful and expensive the electronic device required to implement. Therefore, optimization of cost and accuracy will be considered by the application area the robot will be employed. \par

\subsubsection{Fuzzification}
\textbf{Target Distance membership:}
\begin{figure}[H]
\centering
\includegraphics[width=0.8\textwidth]{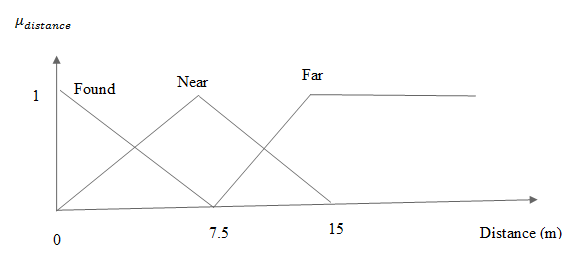}
\caption{Distance linguistic value membership function}
\end{figure}
 \textbf{Target Direction Membership:}
\begin{figure}[H]
\centering
\includegraphics[width=0.8\textwidth]{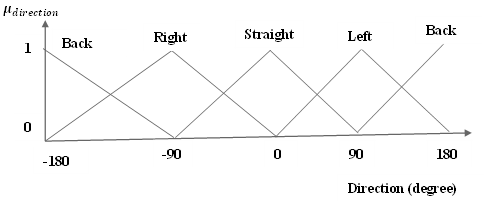} 
\caption{Direction linguistic value membership function}
\end{figure}
\textbf{Obstacle Membership:}
\begin{figure}[H]
\centering
\includegraphics[width=0.8\textwidth]{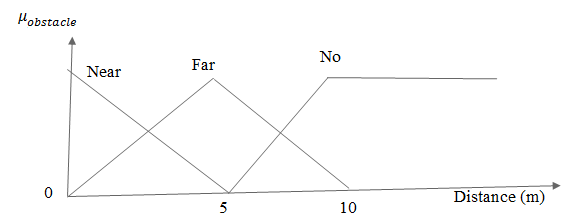} 
\caption{Obstacle membership function}
\end{figure}
\textbf{Robot Wheel Speed Membership:}
\begin{figure}[H]
\centering
\includegraphics[width = 0.8\textwidth]{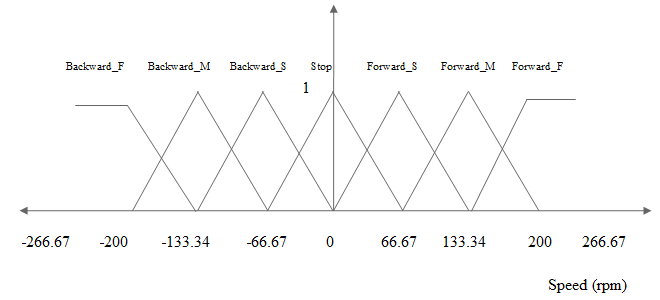} 
\caption{Robot wheel speed membership function}
\end{figure}
\subsubsection{Rule Base }
The rule base is the main part of fuzzy controller design. All inputs fetched from sensors are fuzzified and used to develop the rules. In developing the rule base, there are criteria to be considered. The first one is completeness; the rules should include all possible conditions of the inputs, in other words, there should be an output for all combinations of input states. The other criterion is, that the rules shouldn’t have conflicting decisions. For specific input conditions, there is only one specific decision. Different input conditions could have the same decision but the reverse is not true. Sample rules are displayed in the table below.
\begin{figure}[H]
\centering
\includegraphics[width=0.8\textwidth]{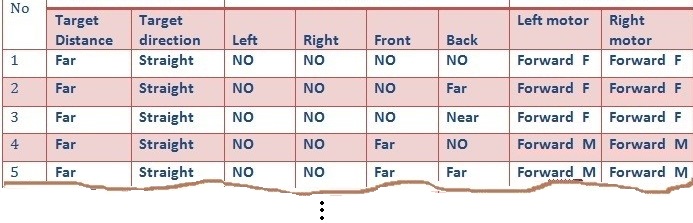} 
\caption{Sample Rule Base: (The letters after the underscore represent as follows: S stands for   Slow, M stands for Medium, and F stands for Fast)}
\end{figure}

\subsubsection{Fuzzy Inference Mechanism}
The fuzzy inference system is the expertise part of the fuzzy controller which has the capability of both simulating human decision-making based on fuzzy concepts and inferring fuzzy control actions by using fuzzy implications and fuzzy logic rules of inference. In other words, once all the monitored input variables are transformed into their respective linguistic variables (by fuzzification), the inference engine evaluates the set of if-then rules (given in the rule base) and thus a result is obtained which is again a linguistic value for the linguistic variable.
\begin{figure}[H]
\centering
\includegraphics[width=0.8\textwidth5]{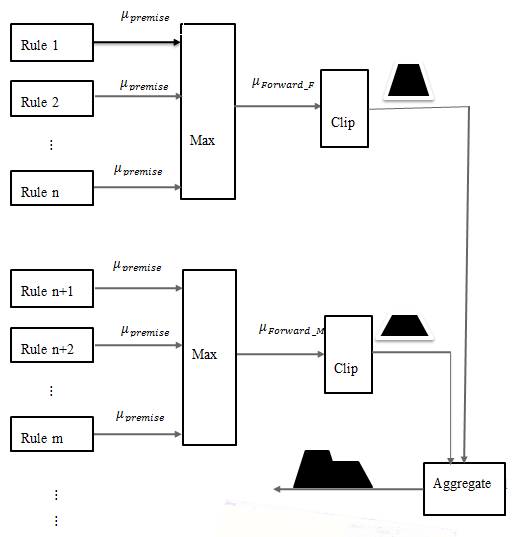} 
\caption{Fuzzy inference mechanism}
\end{figure}
\subsubsection{Defuzzification}
The linguistic result has to be then transformed into a crisp output value of the fuzzy logic controller and this is why there is a second transformation in the fuzzy logic controller. The second transformation is performed by the defuzzifier which performs scale mapping as well as defuzzification. The fuzzified yields a non-fuzzy, crisp control action from the inferred fuzzy control action by using the consequent membership functions of the rules.
\begin{figure}[H]
\centering
\includegraphics[width=1.1\textwidth]{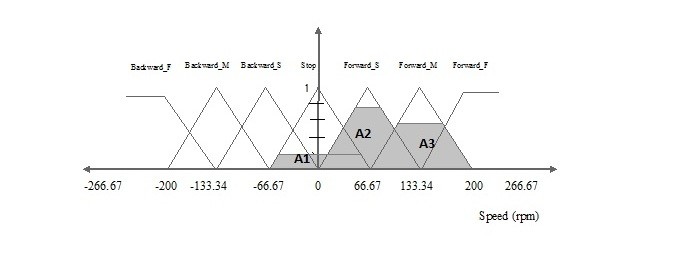} 
\caption{Defuzzification by centroid method}
\end{figure}
The most popular defuzzification method is centroid calculation, which returns the center of the area under the curve. 
\begin{equation}
Speed_{crisp} = \frac{\sum_{i=1}^{7}{A_iW_i}}{\sum_{i=1}^{7}{A_i}}
\end{equation}
where $A_i$ is the area of the clipped triangle (shaded trapezium), $W_i$ is the middle angular speed corresponding to the area $Speed_{crisp}$ is the output of the defuzzification process which is the speed of the motor recommended by the fuzzy controller.

\subsubsection{Control System Structure}
The complete block diagram of the mobile robot navigation control system is shown as follows. 
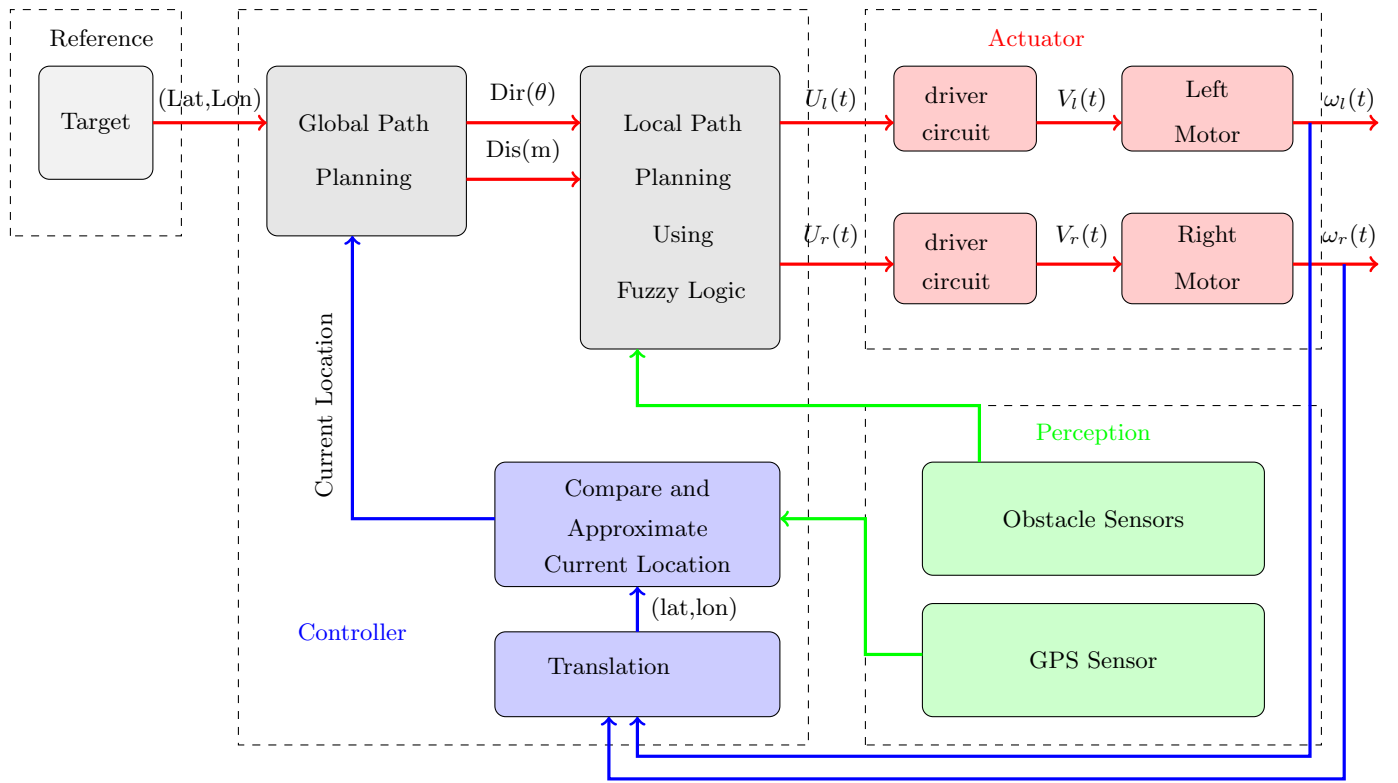
\begin{figure}[H]
\centering
\begin{tikzpicture}[scale=1.5]
\draw [dashed](-5.5,4)--(-5.5,6)--(-4,6)--(-4,4)--(-5.5,4);
\draw [dashed](-3.5,6)--(1.5,6)--(1.5,-0.5)--(-3.5,-0.5)--(-3.5,6);
\draw [dashed](2,3.0)--(2,6)--(6,6)--(6,3.0)--(2,3.0);
\draw [dashed](2,-0.5)--(2,2.5)--(6,2.5)--(6,-0.5)--(2,-0.5);

\draw [rounded corners,fill=gray!10](-5.25,4.5)rectangle(-4.25,5.5);
\draw (-4.75,5) node {\small{Target}};

\draw [rounded corners,fill=gray!20](-3.25,4) rectangle (-1.5,5.5);
\draw (-2.4,5)node{\small{Global Path}};
\draw (-2.4,4.5)node{\small{Planning}};
\draw [rounded corners,fill=gray!20](-0.5,3) rectangle (1.25,5.5);
\draw (0.4,5)node{\small{Local Path}};
\draw (0.4,4.5)node{\small{Planning}};
\draw (0.4,4)node{\small{Using }};
\draw (0.35,3.5)node{\small{ Fuzzy Logic}};
\draw [rounded corners,fill=blue!20](-1.25,-0.25)rectangle (1.25,0.5);
\draw (-0.25,0.2)node{\small{Translation}};
\draw [rounded corners,fill=blue!20](-1.25,0.9)rectangle (1.25,2);
\draw (0,1.75)node{\small{Compare and}};
\draw (0,1.4)node{\small{Approximate}};
\draw (0,1.1)node{\small{Current Location}};

\draw [rounded corners,fill=red!20](2.25,4.75) rectangle (3.5,5.5);
\draw (2.8,5.25)node {\small{driver }};
\draw (2.8,4.92)node {\small{circuit}};
\draw [rounded corners,fill=red!20](4.25,4.75) rectangle (5.75,5.5);
\draw (5.0,5.3)node {\small{Left}};
\draw (5.0,4.9)node {\small{Motor}};
\draw [rounded corners,fill=red!20](2.25,3.4) rectangle (3.5,4.2);
\draw (2.8,3.95)node {\small{driver }};
\draw (2.8,3.6)node {\small{circuit}};
\draw [rounded corners,fill=red!20](4.25,3.4) rectangle (5.75,4.2);
\draw (5.0,4)node {\small{Right}};
\draw (5.0,3.6)node {\small{Motor}};

\draw [rounded corners,fill=green!20](2.5,1.0)rectangle (5.5,2);
\draw (4,1.5)node {\small{Obstacle Sensors}};
\draw [rounded corners,fill=green!20](2.5,-0.25)rectangle (5.5,0.75);
\draw (4,0.25)node {\small{GPS Sensor}};

\draw [->,very thick,color=red](-4.25,5)--(-3.25,5);
\draw (-3.75,5.2)node {\small{(Lat,Lon)}};
\draw [->,very thick,color=red](-1.5,5)--(-0.5,5);
\draw (-1,5.25)node {\small{Dir($\theta$)}};
\draw [->,very thick,color=red](-1.5,4.5)--(-0.5,4.5);
\draw (-1,4.75)node {\small{Dis(m)}};

\draw[->,very thick,color=red](1.25,5)--(2.25,5);
\draw (1.7,5.2)node {\small{$U_l(t)$}};
\draw[->,very thick,color=red](1.25,3.75)--(2.25,3.75);
\draw (1.7,4)node {\small{$U_r(t)$}};

\draw[->,very thick,color=red](3.5,5)--(4.25,5);
\draw (3.9,5.2)node {\small{$V_l(t)$}};
\draw[->,very thick,color=red](3.5,3.75)--(4.25,3.75);
\draw (3.9,4)node {\small{$V_r(t)$}};

\draw[->,very thick,color=red](5.75,5)--(6.5,5);
\draw (6.25,5.2)node {\small{$\omega_l(t)$}};
\draw[->,very thick,color=red](5.75,3.75)--(6.5,3.75);
\draw (6.25,4)node {\small{$\omega_r(t)$}};

\draw [->,very thick,color=blue](5.9,5)--(5.9,-0.6)--(0,-0.6)--(0,-0.25);
\draw [->,very thick,color=blue](6.2,3.75)--(6.2,-0.8)--(-0.25,-0.8)--(-0.25,-0.25);
\draw [->,very thick,color=blue](0,0.5)--(0,0.9);
\draw (0.5,0.7)node{\small{(lat,lon)}};
\draw [->,very thick,color=blue](-1.25,1.5)--(-2.5,1.5)--(-2.5,4);
\draw (-2.75,2.5)node[rotate=90]{\small{Current Location}};
\draw [->,very thick,color=green](2.5,0.3)--(2.0,0.3)--(2.0,1.5)--(1.25,1.5);
\draw [->,very thick,color=green](3,2)--(3,2.5)--(0,2.5)--(0,3);

\draw (-4.7,5.75) node {\small{Reference}};
\draw (-2.5,0.5) node [color=blue]{\small{Controller}};
\draw (3.5,5.75) node [color=red]{\small{Actuator}};
\draw (4,2.25) node [color=green]{\small{Perception}};
\end{tikzpicture}
\caption{Mobile Robot Control System Structure}
\end{figure}
\section{Simulation Results }
In this section, the various results obtained during the research are presented and discussed in detail. The dynamically changing environment will be simulated using MATLAB software and the response of the controller due to the changing environment will be monitored with the help of graphs and 3-D animation. 
This simulation uses the following data (Robot mass = 6Kg, Length between rear wheels = 32 cm, Wheel radius = 7cm, and motor constant = 0.1 V/ (rad/s)) Friction is ignored.
\begin{flushleft}
\begin{figure}[H]
\centering
\includegraphics[width=0.8\textwidth]{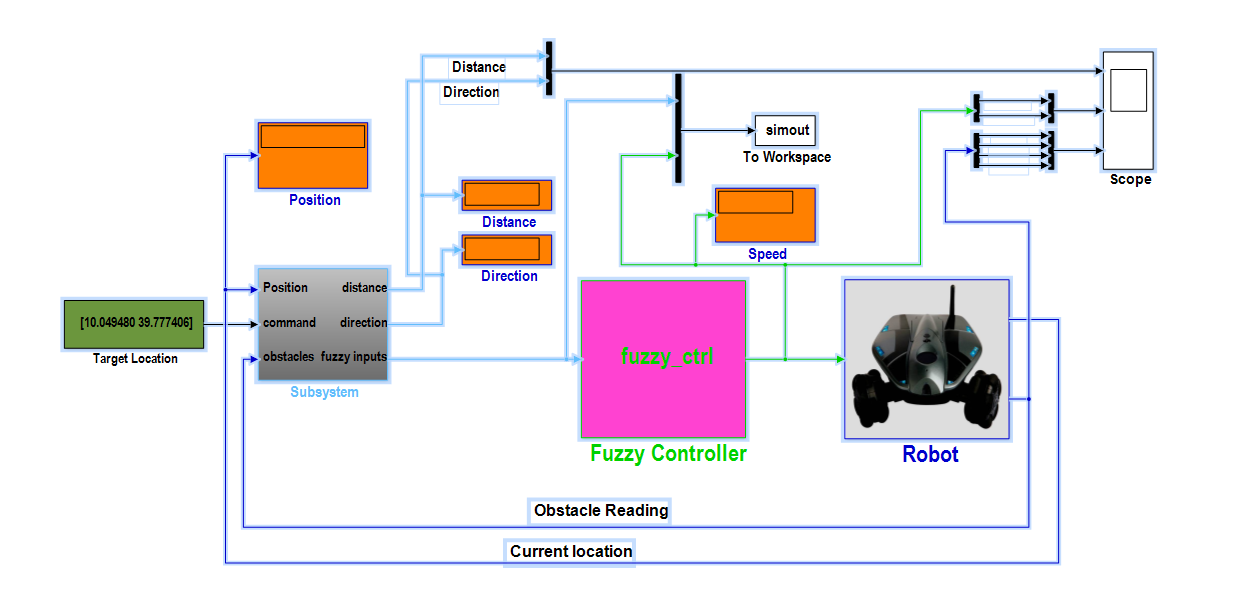} 
\caption{Robot Control System MATLAB model}
\end{figure}
\end{flushleft}
\subsection{Results}
The fuzzy controller performance is evaluated in the MATLAB Simulink model using 3-D animation. The following snap shot is taken from MATLAB simulation.
\begin{figure}[H]
\centering
\includegraphics[width=0.8\textwidth]{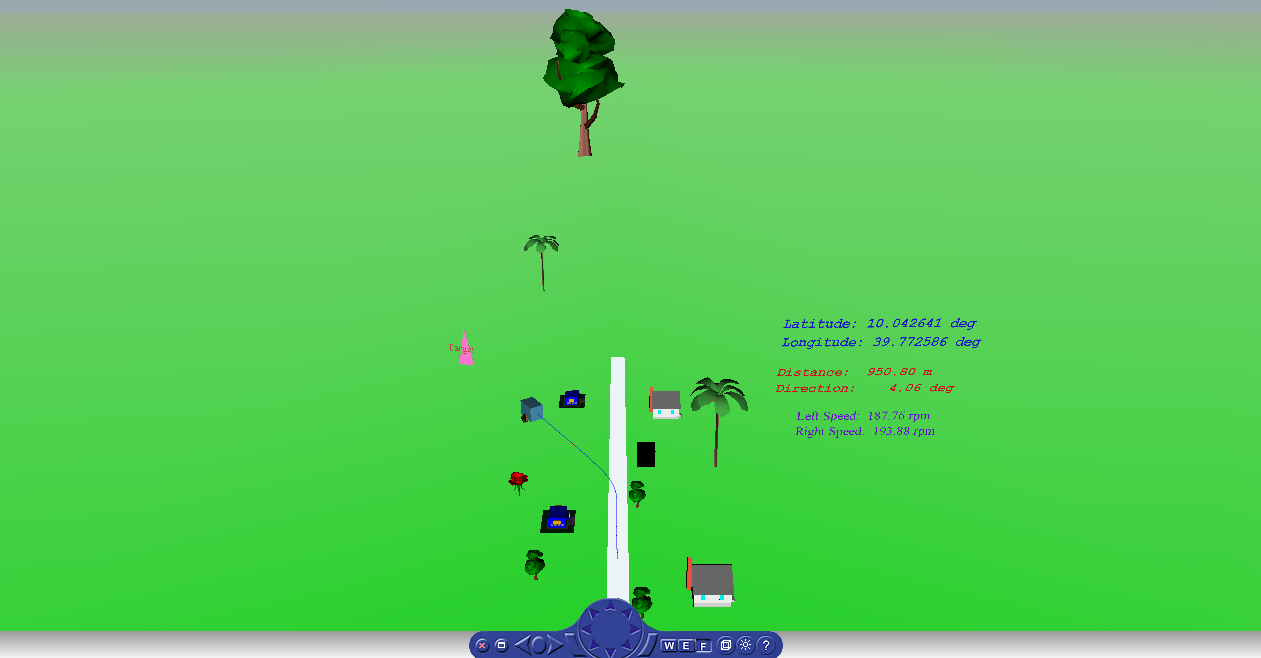} 
\caption{3-D Environment model}
\end{figure}
There are six numbers of input and two output variables that should be monitored for mobile robot navigation. Surrounding information (obstacles and location statuses) is detected using different sensors. Based on all these input conditions the controller will make a decision on which direction and at what speed the robot should move. All these real-time conditions are represented in MATLAB to see the effect of one on another graphically. The following graph shows the response of the fuzzy controller on the speed of two right and left motors with the changing of front obstacles and the location of the mobile robot. 
\vspace{-2cm}
\begin{figure}[H]
\centering
\label{fig77}
\includegraphics[width=0.8\textwidth]{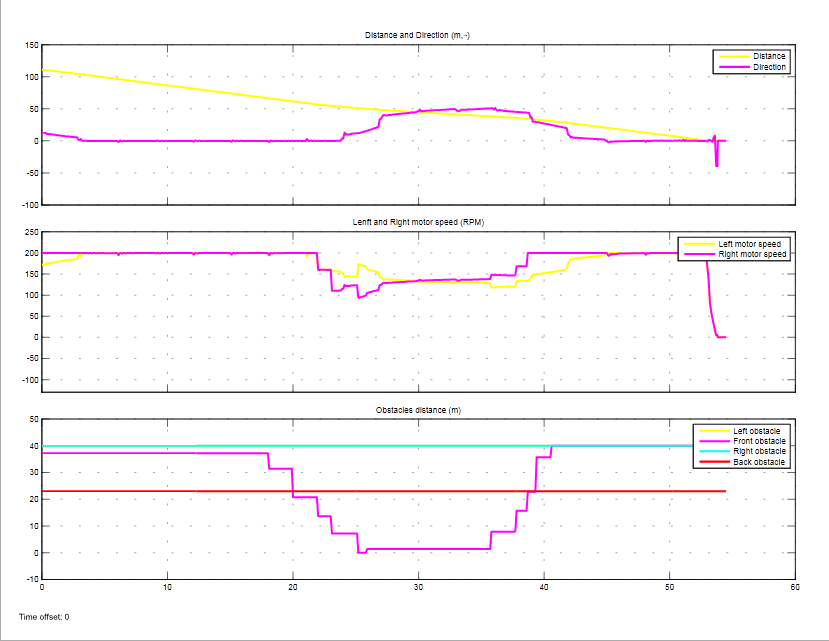} 
\caption{Left and Right motor Speed vs Front obstacle, distance, and direction}
\end{figure}

As we can see in Fig. 17 the robot starts its motion with the right motor speed of \textbf{200 rpm }and the left motor at a speed of \textbf{170 rpm} in forward direction. Then the motion is interrupted at around 20 seconds with the front obstacle approaching. Immediately the controller gives a response by changing the speed of the two back motors as shown. Since there is only one obstacle in the front direction the other 3 sides of the robot are free. So the controllers choose to go in to right direction at a slower speed. When the obstacle is removed, the speed of the motors changes to a new value which will drive the robot to the proper location from the new location. When the robot approaches its target location, the speed of the two motors starts to decrease as we can see on the graph. Distance and direction to the target are also shown on the graph. \par
In Fig. 17 only front obstacles are considered. To see the response of the controller from a different perspective, in Fig. 18 the effect of all input conditions varying randomly is considered. 
\begin{figure}
\centering
\label{fig78}
\includegraphics[width= 0.8\textwidth]{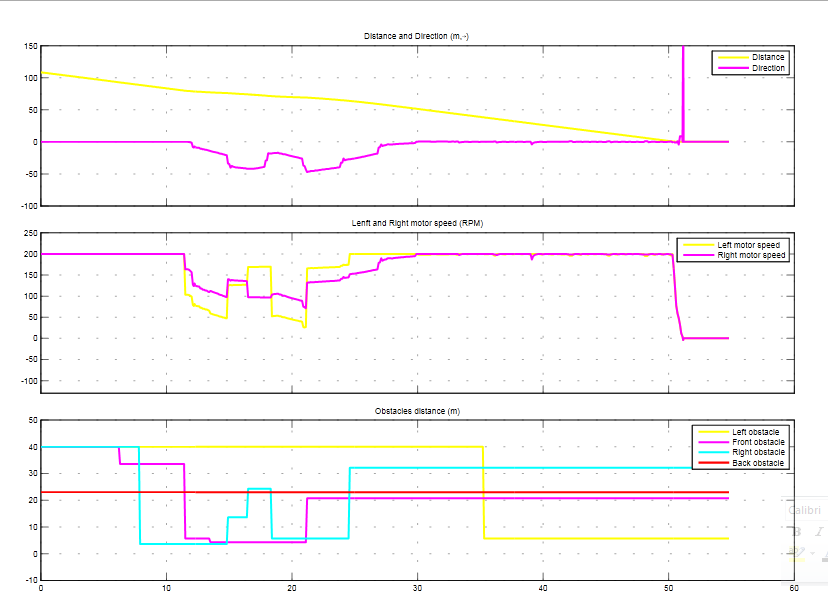} 
\caption{Motor Speed vs. obstacles and location status (obstacles varying randomly)}
\end{figure}
At around 9 seconds obstacle from the right comes near to the robot, but the controller doesn't respond as it is free from the front side.  At around 12 seconds, the front obstacle is set to very near to the robot. Then the controller sets the motor's speed in to differential manner that will drive the robot to avoid the obstacle. When the robot moves around the obstacle the direction of the target also changes, then the controller develops a new path plan from the new location and continues in this way until the target is found. As shown on the graph at 51 seconds the distance to the target becomes zero, and then the controller sets the motors to stop status.\par
We can also see the tracking performance of the fuzzy controller for a given path trajectory. The path planner will generate a path dynamically that will take the robot to the destination by excluding obstacles on its way. Then the robot will follow that path.  Fig. 19 shows us the planned path versus the path followed by the robot.
\begin{figure}[H]
\centering
\label{fig79}
\includegraphics[width=0.8\textwidth]{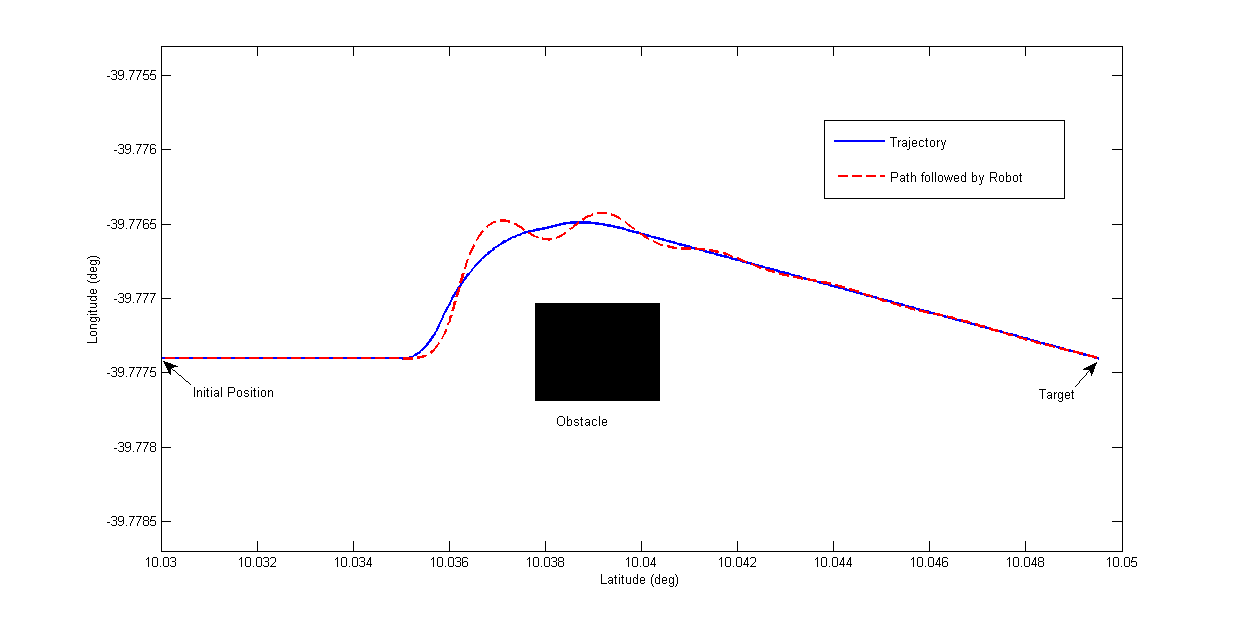} 
\caption{Planned path vs.  Path followed by the robot}
\end{figure}
As indicated in Fig. 19, the controller has difficulty following the exact path line when the path direction changes very quickly. The error is shown in  Fig. 20.
\begin{figure}[H]
\centering
\label{fig80}
\includegraphics[width=0.8\textwidth]{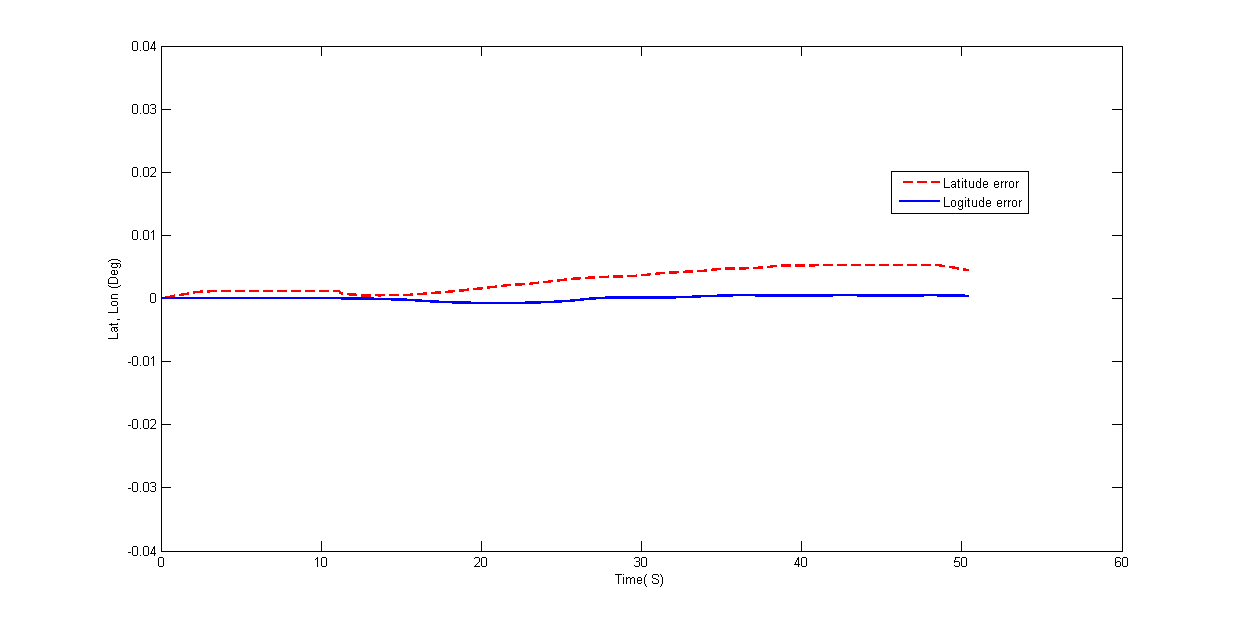} 
\caption{Tracking error}
\end{figure}

\section{Conclusion and Future Work}
\subsection{Conclusion}
This research covered mobile robot motion controller design and simulation. Mobile robot navigation to goal position in an unknown dynamic outdoor environment is identified to be a non-linear problem with multiple control variables. The path planning and obstacle avoidance problem is addressed by using a fuzzy logic control algorithm. One of the advantages of the use of fuzzy logic control resides in the possibility of averaging the conclusion of several rules to obtain an intermediary action. The basic idea behind a fuzzy logic controller is to avoid the design of a strategy based on a detailed dynamic model, by taking the approach of a human operator to an ill-defined system.\par
Simulations were carried out on differential drive non-holonomic mobile robots to test the performance of the proposed fuzzy controller. The mobile robot was seen to reach the target location efficiently with obstacle conditions.\par
However, the modeling of a fuzzy controller is an iterative process of trial and error. It involves tuning the shapes of the membership function as well as fuzzy rule bases. For relatively complex control problems this synthesis procedure can turn out to be lengthy depending on the availability of expert knowledge or intuition. This study proposes the use of optimization techniques like Genetic Algorithms and Neural Networks to tune fuzzy membership functions and rule bases for better results.\par
The obstacle avoidance fuzzy algorithm performs nicely with the limitation of the inability to identify passable narrow paths. This problem is caused by the limitation of sonar sensors to measure the distance between obstacles and the size of obstacles. This can be solved by using camera or 3-D sensors.  \par

Overall it became clear that fuzzy logic is a suitable control system where the control of mobile robots is concerned. The simulation result shows that the fuzzy controller was successful at following a defined path and avoiding obstacles.



\begin{thebibliography}{}
\bibitem{1}Roland Illah, R. Siegwart  Nourbakhsh: \emph{Introduction to Autonomous  Mobile  Robot}, Massachusetts Institute of Technology, 2004
\bibitem{2}	Kevin M. Passino and Stephen Yurkovich,\emph{ Fuzzy Control}, The Ohio State University, 1998.
\bibitem{3}	Dr. Kevin Craig, \emph{Fuzzy Logic and Fuzzy Control}, Rensselaer Polytechnic Institute
\bibitem{4}	\emph{Fuzzy Logic $Toolbox^{TM}$ 2 User’s Guide}, $Mathworks^{TM}$,2008
\bibitem{5}	Brian Hahn. Daniel T. Valentine, \emph{Essential MATLAB for Engineers and Scientists}, Third Edition, 2007
\bibitem{6}	SIMULINK,\emph{ Dynamic System Simulation for MATLAB}, January 1999
\bibitem{7}	\emph{Virtual Reality Toolbox, For Use with MATLAB and SIMULINK}
\bibitem{8}	Stanislaw H.  Zak,\emph{ Systems and Controls}, Oxford University Press, 2003
\bibitem{9}	 Vamsi Mohan Peri, Dan Simon, \emph{Fuzzy Logic Control for an Autonomous Robot},   Cleveland State University Cleveland,Ohio,USA.
\bibitem{10}Krzysztof Kozlowski, Dariusz Pazderski , \emph{Modeling and Control of A 4-Wheel Skid-Steering Mobile Robot}, Poznan University of Technology, 2004
\bibitem{11} Maria Isabel Ribeiro Pedro Lima, \emph{Kinematics Models of Mobile Robots}, Portugal,April 2002
\bibitem{12}Gyula Mester , Aleksandar Rodić,\emph{ Sensor-Based Intelligent Mobile Robot Navigation in Unknown Environments},  University of Szeged, Department of Informatics, Volume 1, Number 2, November 2010
\bibitem{13}Tim Bailey, \emph{Mobile Robot Localization and Mapping in Extensive Outdoor Environment}, The University of Sydney, August 2002
\bibitem{14}Christopher Peter Lee-Johnson,\emph{ The Development of a Control System for an Autonomous Mobile Robot}, University of Waikato, 2004
\bibitem{15}Vamsi Mohan Peri, \emph{Fuzzy Logic Controller for an Autonomous Mobile Robot}, Jawaharlal Nehru Technological University, India, May, 2002
\bibitem{16}Andre Michael Swartz, \emph{Methods for Designing and Optimizing Fuzzy Controllers}, December 1999
\bibitem{17}Stergios I. Roumeliotis, \emph{ Robust Mobile Robot Localization: From Single-Robot Uncertainties to Multi-Robot Interdependencies}, University of Southern California,  May 2000
\bibitem{18}Codrin Pasca, \emph{History of Robotics, Robotics – Intelligent Connection of the Perception to Action}, May 5, 2003.
\bibitem{19}Ayman AbuBaker, \emph{A Novel Mobile Robot Navigation System Using Neuro-Fuzzy Rule-Based Optimization Technique}, 2012.
\bibitem{20}Jia-Heng Zhou and Huei-Yung Lin, \emph{A Self-Localization and Path Planning Technique for Mobile Robot Navigation}, 2001.
\bibitem{21}Peter Corke, \emph{Robotics,Vision and Control}, 2011.
\bibitem{22}Prases K. Mohanty* and Dayal R. Parhi,\emph{ Controlling the Motion of an Autonomous Mobile Robot Using Various Techniques}, 2012.
\bibitem{23}J. Giesbrecht , \emph{Global Path Planning for Unmanned Ground Vehicles}, DEC 2004
\bibitem{24}Y. Ono, H. Uchiyama, W. Potter, \emph{A Mobile Robot for Corridor Navigation:  A Multi-Agent Approach}, 2004
\bibitem{25}Hani Hagras, \emph{A Type-2 Fuzzy Logic Controller For Autonomous Mobile Robots}, 2004
\bibitem{26}G.N. Marichal, L. Acosta, L. Moreno, J.A. M-endez, J.J. Rodrigo, M. Sigut, \emph{Obstacle avoidance for a mobile robot: A neuro-fuzzy approach}, 30 May 2000
\bibitem{27}J.L. Fernández, R. Sanz and A.R. Diéguez, \emph{The Beam-Curvature Method: A New Approach For Improving Local Realtime Obstacle Avoidance}, 2002
\bibitem{28}Alejandro Ramirez-Serrano and Marc Boumedine,\emph{ Real Time Navigation in Unknown Environments Using Fuzzy Logic and Ultrasonic Sensors}, IEEE, 1996
\bibitem{29}B.N. Kimt,  Kwont, K.H. Kimtt, E.H. Leet, S.H. Hong,\emph{ A Study On Path Planning For Mobile Robot Based On Fuzzy Logic Controller}, IEEE, 1999
\bibitem{30}M. Norouzi, A. Karambakhsh, M. Namazifar and B. Savkovic, \emph{Object Based Navigation of Mobile Robot with Obstacle Avoidance using Fuzzy Controller}, IEEE, 2009
\bibitem{31}S.M.Raguraman, D.Tamilselvi and N.Shivakumar, \emph{Mobile Robot Navigation Using Fuzzy logic Controller}, 2009
\bibitem{32}Limin Ren1, Weidong Wang  and Zhijiang Du, \emph{A New Fuzzy Intelligent Obstacle Avoidance Control Strategy for Wheeled Mobile Robot}, IEEE, 2012
\bibitem{33}Mohannad Al-Khatib, \emph{Performance Measures And Preferences For Fuzzy Controllers Design Algorithms}, American University of Beirut, 2004
\end{thebibliography}
\end{document}